\documentclass[runningheads]{llncs}
\usepackage{graphicx}
\usepackage{amsmath}

\begin{document}
\title{Efficient Knowledge Distillation: Empowering Small Language Models with Teacher Model Insights}
\titlerunning{Efficient Knowledge Distillation}
%

\author{
Mohamad Ballout\inst{1} \and
Ulf Krumnack\inst{1} \and
Gunther Heidemann\inst{1} \and 
Kai-Uwe Kühnberger\inst{1}}

\institute{Institute of Cognitive Science, Osnabrueck University, 49074 Osnabrück, Germany \\
\email{mohamad.ballout@uni-osnabrueck.de}}

\authorrunning{Ballout et. al}
%
%
\maketitle              
\begin{abstract}

Enhancing small language models for real-life application deployment is a significant challenge facing the research community. Due to the difficulties and costs of using large language models, researchers are seeking ways to effectively deploy task-specific small models. In this work, we introduce a simple yet effective knowledge distillation method to improve the performance of small language models. Our approach utilizes a teacher model with approximately 3 billion parameters to identify the most influential tokens in its decision-making process. These tokens are extracted from the input based on their attribution scores relative to the output, using methods like saliency maps. These important tokens are then provided as rationales to a student model, aiming to distill the knowledge of the teacher model. This method has proven to be effective, as demonstrated by testing it on four diverse datasets, where it shows improvement over both standard fine-tuning methods and state-of-the-art knowledge distillation models. Furthermore, we explore explanations of the success of the model by analyzing the important tokens extracted from the teacher model. Our findings reveal that in 68\% of cases, specifically in datasets where labels are part of the answer, such as multiple-choice questions, the extracted tokens are part of the ground truth.

\end{abstract}
\section{Introduction}

Large language models are increasingly becoming integrated into our daily lives, with applications now extending to small devices such as phones and computers. However, the main challenge in this rapid development is the computational power required to run these large language models. The current state-of-the-art models, which have hundreds of billion of parameters, require hundreds of gigabytes of memory for operation. This level of computational power is only affordable for a limited number of research teams. This disparity is evident in the download numbers of smaller models versus larger models. For example, despite Llama2 \cite{lama2} being a powerful open-source state-of-the-art model released in 2023, its downloads on Hugging Face \cite{hf} are significantly (around 10 times) lower than those of T5-base \cite{t5}, a smaller language model released in 2019. This discrepancy has encouraged researchers to develop new methods for distilling knowledge from large models into smaller ones \cite{distillation,distillation2,distillation3,step-step}, making them more feasible for integration on portable devices and accessible to a broader range of practitioners.

Knowledge distillation is a technique that leverages the knowledge of a larger, more complex model (the teacher) to train a smaller, more efficient model (the student) through teacher-student prediction alignment. One effective method within this framework is gradient-based knowledge distillation \cite{grad-distillation1,grad-distillation2}. In this approach, the gradient, which describes how the teacher model's predictions change in response to variations in inputs, is used as a guide for the student model. This guidance is beneficial for the student model to more accurately approximate the teacher model's underlying functionality and decisions, thereby enhancing the student's learning and performance.

Another proven and effective practice for enhancing the performance of large language models (LLMs) involves providing them with rationales. This enhancement can be implemented in two main ways. For larger models, particularly those with over 100 billion parameters, rationales can be incorporated by including step-by-step instructions in the prompt. Techniques such as the ``Chain-of-Thought'' \cite{cot} or ``Tree-of-Thought"  \cite{tot} methods are examples of this strategy. In contrast, smaller models are typically improved through fine-tuning with instructional data \cite{instruction1,instruction2,instruction3}, as prompting smaller models with step-by-step instructions has not yet been demonstrated to be effective \cite{cot}. This fine-tuning approach allows smaller models to effectively integrate structured guidance, thereby enhancing their problem-solving capabilities and overall performance.

In this work, we introduce a novel yet straightforward and effective technique that combines gradient knowledge distillation with the provision of rationales to large language models (LLMs). Our proposed method involves fine-tuning a teacher model, approximately 3 billion parameters in size, and using this model to identify and extract the top-k important tokens from the input. These important tokens are then provided as rationales to the student model. The key tokens are determined using their gradient attribution, where we calculate the gradient of the model's prediction with respect to each input token based on the saliency  map method \cite{saliency}. In this method, the gradient value indicates the sensitivity of the model's prediction to changes in a specific token. A high gradient value for a token in the input suggests that minor changes in that token can lead to significant changes in the model's prediction, highlighting its importance in influencing the model's output. Conversely, a low gradient value implies that the token has a minimal impact on the prediction, indicating its lesser importance. After identifying the important tokens with high attribution values in the teacher model, we then feed these tokens to the student model as rationales, thereby enhancing its learning process and performance.

\begin{figure}[t]
\includegraphics[scale=0.3]{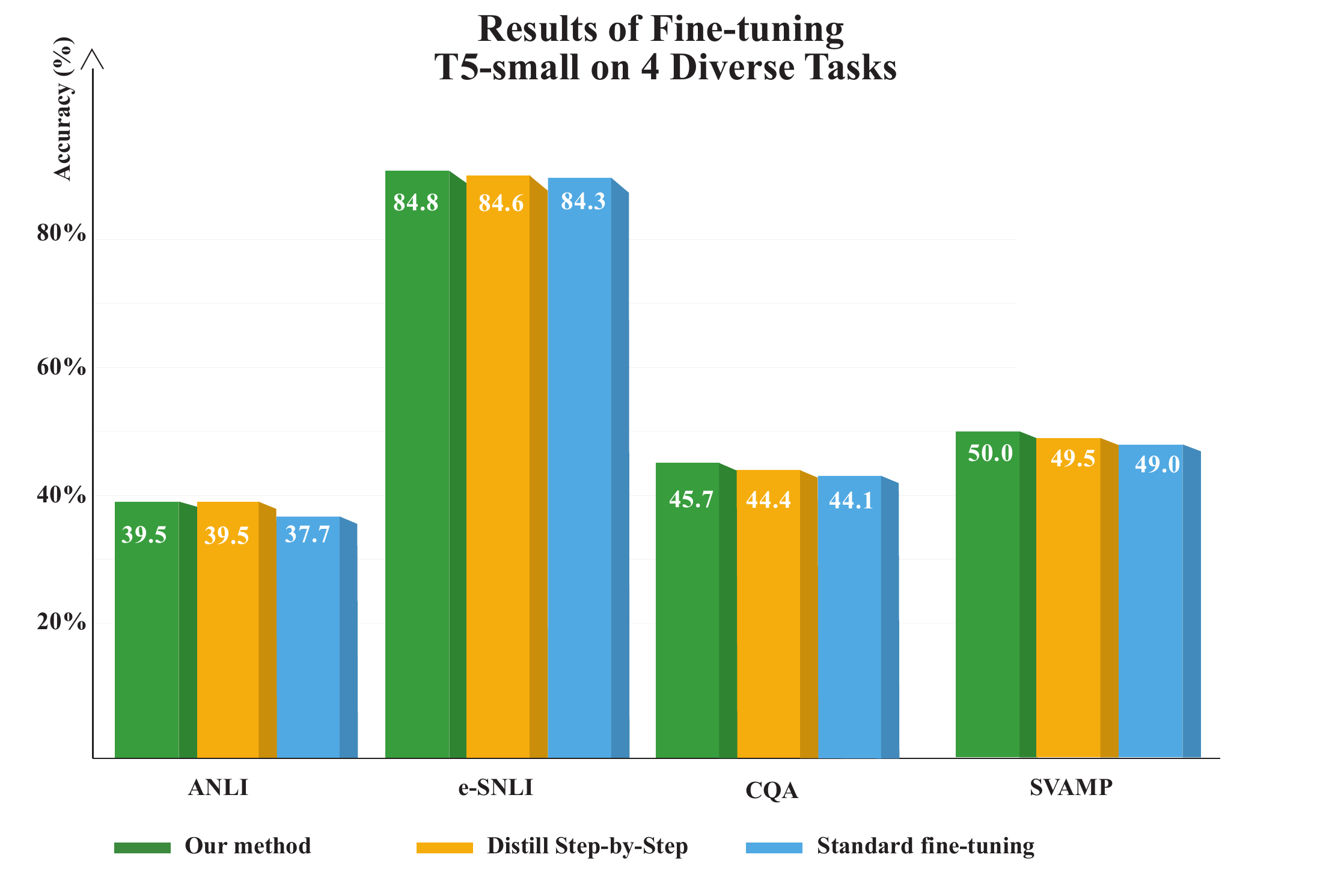}
\caption{The figure displays the performance of our proposed method across four diverse datasets: e-SNLI \cite{esnli} and ANLI \cite{anli}, which involve natural language inference; CQA \cite{cqa1,cqa2} focusing on commonsense question answering; and SVAMP \cite{svamp}, which deals with arithmetic word problems. We compare our method to both the standard fine-tuning method and the step-by-step distillation method.} \label{fig1}
\end{figure}

In order to exemplify the procedure, consider an input from a dataset: ``A boss may like an employee's ambition, so the employee may get put what? Choices are (a) in charge of project (b) conquer opponent (c) go to school (d) begin work (e) webisode." In our approach, we first calculate the attribution of each word in this input using the teacher model. Let's say the significant words identified are ``boss," ``employee," ``charge," ``ambition," and ``what." Next, we fine-tune the student model not only to predict the correct answer but also to generate these key words as rationales. Consequently, the student model's output should be: ``boss, employee, charge, ambition, what, so the answer is (a) in charge of project." The loss is calculated based on this entire output, while the accuracy is assessed solely based on the correctness of the answer, which in this case is ``in charge of project." We tested this method on four diverse datasets and observed improvements compared to standard fine-tuning, as well as a distillation method that uses a very large model (540 billion parameters)  \cite{step-step} as demonstrated in figure \ref{fig1}.   

\section{Related Work}

Due to the importance and the need to improve the performance of the smaller models, knowledge distillation is a hot topic that is grabbing plenty of attention recently. Another hot topic that is related to our work is providing rationales to LLMs in order to improve their performance. Thus this section is dedicated to review the work on these two topics.

\subsubsection{Rationales.}

The use of rationales with language models started with large models \cite{cot,tot,tot2,tot3,tot4} and eventually extended to smaller language models \cite{instruction1,instruction3,instruction4,instruction5}. With large models methods like ``Chain-of-thought" \cite{cot,cot2} gained popularity due to its effectiveness and its similarity of the human cognitive process. Instead of providing only the problem and its solution, the models are provided with few human examples of the intermediate steps that lead the model to predict the answer. The problem faced when using the same technique with smaller models is that they are not few shot learner, and generating human labeled step-by-step instruction is costly. Thus, a work around this problem is to make the large language model to generate step-by-step instructions for each sample in the dataset by providing the large model with few-shot examples on how to generate these steps. For example, \cite{step-step} generated instructions for 4 different dataset using 540B PaLM  model \cite{palm} after providing it with few examples on how to generate it. The benefits of our method over theirs is that first our teacher model is only 3 billion parameters compared to 540 billion parameters. Second, we did not need any human labeling to generate the rationales in contrast of theirs. With these two benefits in mind, our model was still able to perform better with the student model on multiple datasets.

\subsubsection{Gradient-based Knowledge Distillation.}

Knowledge distillation \cite{distillation,gkd} is a successful method to transfer knowledge from a teacher model to a smaller model. It has been used with computer vision \cite{distillation2,vision-distil1,vision-distil2,vision-distil3} and in NLP \cite{step-step,step-step2,step-step3,flan} and other deep learning applications. While attribution analysis methods like saliency maps \cite{saliency}, and Integrated Gradients (IG) \cite{ig} are used for interpetibilty of the models \cite{gradient-exp1,gradient-exp2}, few works have explored their use in knowledge distillation in what we can call 
gradient-based knowledge distillation \cite{grad-distillation2}. In this paper \cite{grad-distillation2}, multi-view attribution maps of the teacher model are extracted and then the difference between the two normalized sets of maps in the teacher and student models is minimized using distance metrics like L2 distance. Our method requires less computation and is simpler to apply where we extract only the tokens that are attributing mostly to the output and providing them to the student model as rationales.

\section{Methodology}
\subsection{Our Approach}

\begin{figure}[t]
\centering
\hspace*{-0.3cm} 
\includegraphics[scale=0.78]{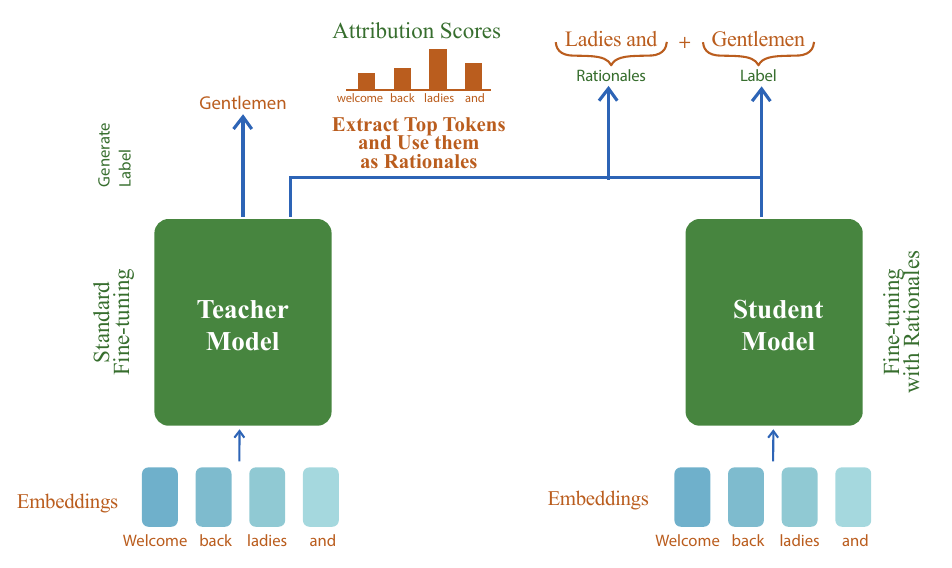}
\caption{The figure illustrates the pipeline of our proposed method, where tokens with the highest attribution scores, extracted from the teacher model, are used as rationales. The student model is then trained to generate these rationales, along with the original labels.} \label{fig3}
\end{figure}

Our method involves providing rationales generated by a teacher model, without human intervention, to a student model. Specifically, we fine-tune a teacher model, T5-flan-3b \cite{flan} in our case, on various datasets. Once the teacher model is fine-tuned, we identify the tokens that contribute most significantly to the output using the saliency map approach, a gradient-based attribution method. This involves computing the gradient of the model's output logits with respect to the input embeddings. We carry out a forward pass using the input embeddings and calculate the gradients for each token in the target sequence. These gradients are then averaged across all tokens in the target sequence. We determine the importance of each input token by computing the average gradient magnitude using the L1 norm. A high average gradient magnitude for a token suggests that it is influential in determining the model's output for the target sequence, while a low magnitude indicates lesser importance. The pipeline of this method is illustrated in figure \ref{fig3}. 

b

Let \( X = \{x_1, x_2, \ldots, x_n\} \) be the input tokens of the sentence, where \( n \) is the number of tokens. The input embeddings are denoted as \( E = \{e_1, e_2, \ldots, e_n\} \), where \( e_i \) is the embedding vector of token \( x_i \).

Let \( f(E) \) be the function representing the forward pass of the model, mapping the input embeddings \( E \) to the output logits. The output logits for the target sequence are given by \( O = f(E) \), where \( O = \{o_1, o_2, \ldots, o_m\} \) and \( m \) is the number of tokens in the target sequence.

The gradient of the output logits with respect to the input embeddings is computed for each token in the target sequence. For each output logit \( o_j \), the gradient with respect to the input embedding \( e_i \) is denoted as \( \nabla_{e_i} o_j \). The gradients are calculated as follows:
\begin{equation}
\nabla_{e_i} O = \{\nabla_{e_i} o_1, \nabla_{e_i} o_2, \ldots, \nabla_{e_i} o_m\}
\end{equation}

The average gradient for each input embedding is computed by averaging across all tokens in the target sequence. The average gradient for embedding \( e_i \) is given by:
\begin{equation}
\bar{G}_i = \frac{1}{m} \sum_{j=1}^{m} \nabla_{e_i} o_j
\end{equation}

The importance of each input token is determined by computing the L1 norm of the average gradient for its embedding. The importance score \( I_i \) for token \( x_i \)is calculated as:
\begin{equation}
I_i = \lVert \bar{G}_i \rVert_1 = \sum_{k} | \bar{G}_i[k] |
\end{equation}

In this equation, \(k\) indexes the elements (or dimensions) of the average gradient vector \(\bar{G}_i\) for a given input embedding \(e_i\). A high value of \( I_i \) implies that the token \( x_i \) is influential in shaping the model's output for the entire target sequence. Conversely, a low value of \( I_i \) suggests that the token \( x_i \) has lesser importance.

After identifying the top-k important tokens with high gradient magnitudes, we provide them to the smaller model to generate them as rationales in addition to the correct answer. We adopt the loss function from \cite{step-step}. In standard fine-tuning, the cross-entropy loss \( \ell \) between the predicted and target tokens, is calculated as shown in equation \ref{eq4} where \( x \) is the input, \( y \) is the output.

\begin{equation}
\mathcal{L}_{\text{label}} = \frac{1}{N} \sum_{i=1}^{N} \ell(f(x_i), {y}_i) \tag{4}
\label{eq4}
\end{equation}

In our study, similar to \cite{step-step}, we approach the integration of rationales into the learning process as a dual-objective task rather than merely incorporating them as extra inputs for the model. Specifically, we train our model, denoted as \( f(x_i) \), to accomplish two goals: predicting the task labels  \( y \) and generating the corresponding rationales \( r \) from the given text inputs. This approach is formalized as the combined loss function as shown in equation \ref{eq5}.
\begin{equation}
\mathcal{L} = \mathcal{L}_{\text{label}} + \lambda \mathcal{L}_{\text{rationale}} \tag{5}
\label{eq5}
\end{equation}

where \( \mathcal{L}_{\text{label}} \) is the loss associated with label prediction as outlined in Equation \ref{eq4}, \( \mathcal{L}_{\text{rationale}} \) represents the loss related to rationale generation, and \( \lambda \) is a hyper-parameter that determines how much the rational loss affect the total loss. In our experiment, we set \( \lambda \) to 0.5. The rationale generation loss, which is expressed in equation \ref{eq6} encourages the model to learn the intermediate reasoning that leads to its predictions. By doing so, it aims to enhance the model's capability by leveraging the generated tokens from the teacher model. \( r_i \) in the equation represents the rationales extracted from the teacher model. 
\begin{equation}
\mathcal{L}_{\text{rationale}} = \frac{1}{N} \sum_{i=1}^{N} \ell(f(x_i), r_i) \tag{6}
\label{eq6}
\end{equation}

\subsection{Datasets}
To compare our model with a state-of-the-art model, we utilized the same datasets as those used in \cite{step-step}. These datasets encompass a range of tasks, including natural language inference, commonsense question answering, and arithmetic word problems.

For natural language inference, we employed the e-SNLI \cite{esnli} and ANLI \cite{anli} datasets. In these datasets, the model is provided with premises and hypotheses and is required to predict whether the hypothesis is an entailment, a contradiction, or neutral with respect to the premise. For example, given the premise ``This church choir sings to the masses as they sing joyous songs from the book at a church" and a hypothesis ``A choir singing at a baseball game.", the model must determine this hypothesis contradicts the premises since the choir is at a church not a baseball game.

Another dataset we used is CQA \cite{cqa1,cqa2}, which focuses on commonsense question answering. Here, the model is presented with an input question accompanied by five answer choices. A typical question might be: ``A person with digestion issues eats a meat-filled breakfast, what does he feel?" with choices like [``heartburn", ``overeating", ``happiness", ``being satisfied", ``gain energy"]. The correct answer in this instance would be ``heartburn."

Lastly, we evaluated our approach using the SVAMP dataset \cite{svamp}, which consists of arithmetic word problems. These problems require the model to generate a numerical answer. An illustrative problem might be: ``Ed had 10 more marbles than Doug. Doug lost 11 of his marbles at the playground. If Ed had 45 marbles, how many more marbles did Ed have than Doug then?". The expected model response would be a calculation like ``10.0 + 11.0'', signifying the numerical solution to the problem.

By testing our model across these diverse datasets, we aimed to demonstrate its versatility and effectiveness in handling various types of language processing tasks.  

\section{Results}
\subsection{Results of Using Extracted Tokens as Rationales}
In our study, we compare our proposed methods with two other approaches: standard fine-tuning, where the model is trained to generate a label from input, and the distillation step-by-step method \cite{step-step}. The distillation step-by-step method utilizes a teacher model named Palm, which has approximately 540 billion parameters. This model is provided with a few human-labeled examples to generate step-by-step instructions. Conversely, our teacher model is Flan-T5-3b, with around 2.7 billion parameters. We employ the Hugging Face \cite{hf} implementation of these models, using a learning rate of 5e-5.

Our approach was tested on two small student models, t5-small and Flan-T5-small, each having around 60 million parameters. To replicate the results of \cite{step-step}, we used their provided code and hyper-parameters, adopting their fine-tuning process for the student models. We chose to extract and use 5 important tokens from the teacher model as rationales for the smaller model. The reason behind using 5 tokens will be discussed in a later section.

The results, as shown in table \ref{tab1}, indicate that our approach outperforms standard fine-tuning on all datasets and also surpasses the distillation step-by-step method on all datasets except ANLI. This comparison highlights the effectiveness of our method, particularly in the context of smaller models where the balance between model size and performance is critical. Our findings suggest that extracting key tokens from a relatively smaller teacher model can significantly enhance the performance of student models across a variety of datasets.     

\begin{table}[htbp]
\caption{The table presents the results of fine-tuning T5-small using three approaches: standard fine-tuning, step-by-step distillation, and our method, across four different datasets.}\label{tab1}
\centering
\begin{tabular}{|c|c|c|c|c|}
\hline
\textbf{T5-small} &   \textbf{ANLI} &  \textbf{e-SNLI} &  \textbf{CQA} &  \textbf{SVAMP} \\
\hline
Standard fine-tuning & 37.7  & 84.3  & 44.1  & 49.0 \\
Dist. step-by-step &  \textbf{39.5}  & 84.6  & 44.4 & 49.5 \\
Ours &  \textbf{39.5}  &  \textbf{84.8}  &  \textbf{45.7}  &  \textbf{50.0}  \\
\hline
\end{tabular}
\end{table}

To verify the versatility of our method with different models, we tested our approach using a more advanced model, FLAN-T5-small. Table \ref{tab2} presents the results of this test, demonstrating the effectiveness of our method compared to the standard fine-tuning approach. When FLAN-T5-small is used as the baseline model, our method shows improvements on all datasets compared to standard fine-tuning, with the exception of SVAMP.

\begin{table}[htbp]
\caption{The table presents the results of fine-tuning Flan-T5-small using three approaches: standard fine-tuning, step-by-step distillation, and our method, across four different datasets.}\label{tab2}
\centering
\begin{tabular}{|c|c|c|c|c|}
\hline
\textbf{Flan-T5-small} &   \textbf{ANLI} &  \textbf{e-SNLI} &  \textbf{CQA} &  \textbf{SVAMP} \\
\hline
Standard fine-tuning & 39.6  & 87.2  & 48.4  & \textbf{53.0} \\
Dist. step-by-step & 40.2  & 87.6  & \textbf{49.3} & \textbf{53.0} \\
Ours &  \textbf{40.3}  &  \textbf{87.8}  & 49.0  &  52.0  \\
\hline
\end{tabular}
\end{table}

\subsection{The effect of Number of Words}
To analyze the impact of the number of ``important" words used as rationales in the student model, we conducted fine-tuning experiments on the T5-small model using a range of values for ``k", where ``k" represents the number of important words extracted from the teacher model. These experiments were performed on the ANLI and CQA dataset, with ``k" varying from 1 to 10.

\begin{figure}[htbp]
\centering
\begin{minipage}{.5\textwidth}
  \centering
  \includegraphics[scale=0.6]{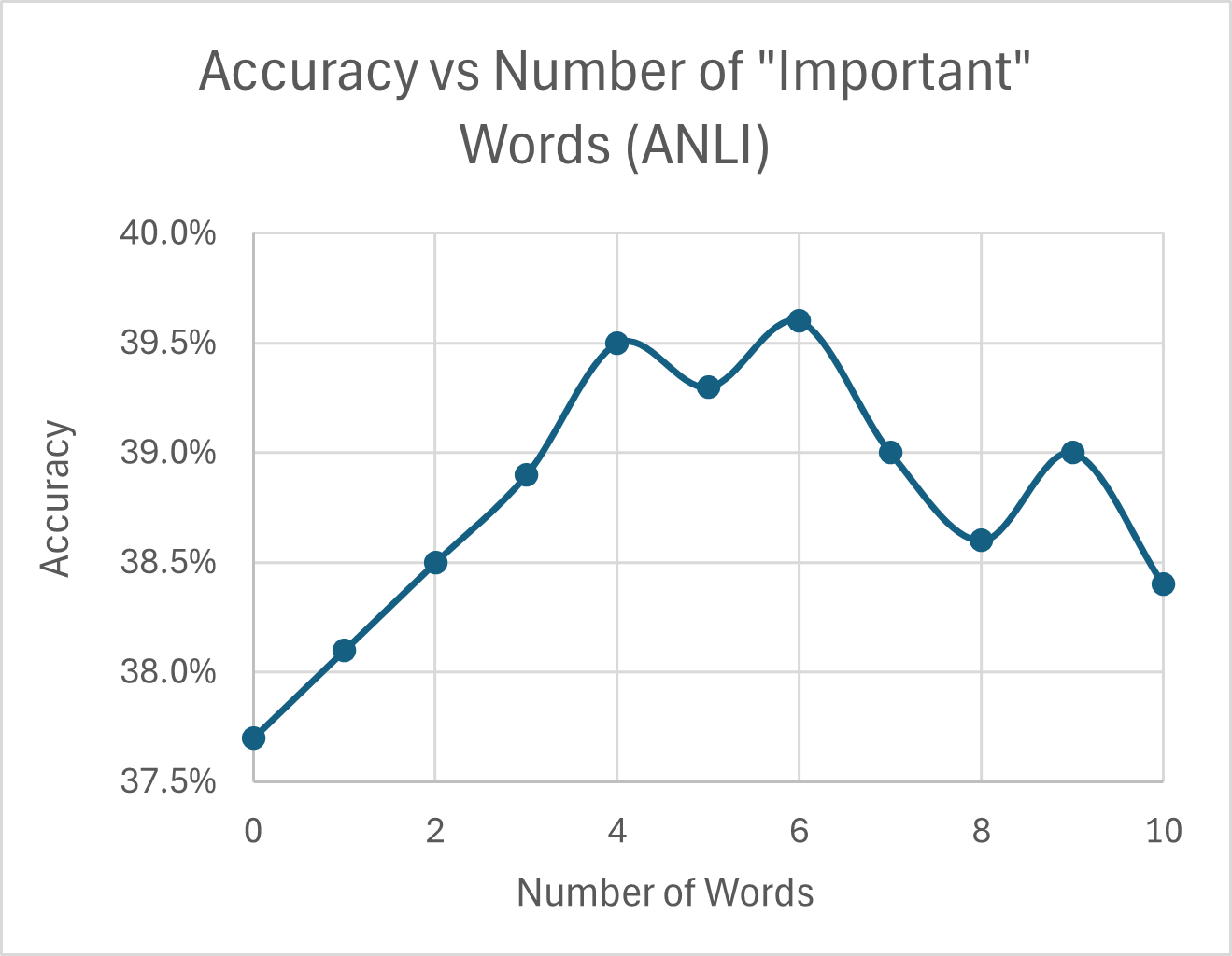}

\end{minipage}%
\begin{minipage}{.5\textwidth}
  \centering
  \includegraphics[scale=0.6]{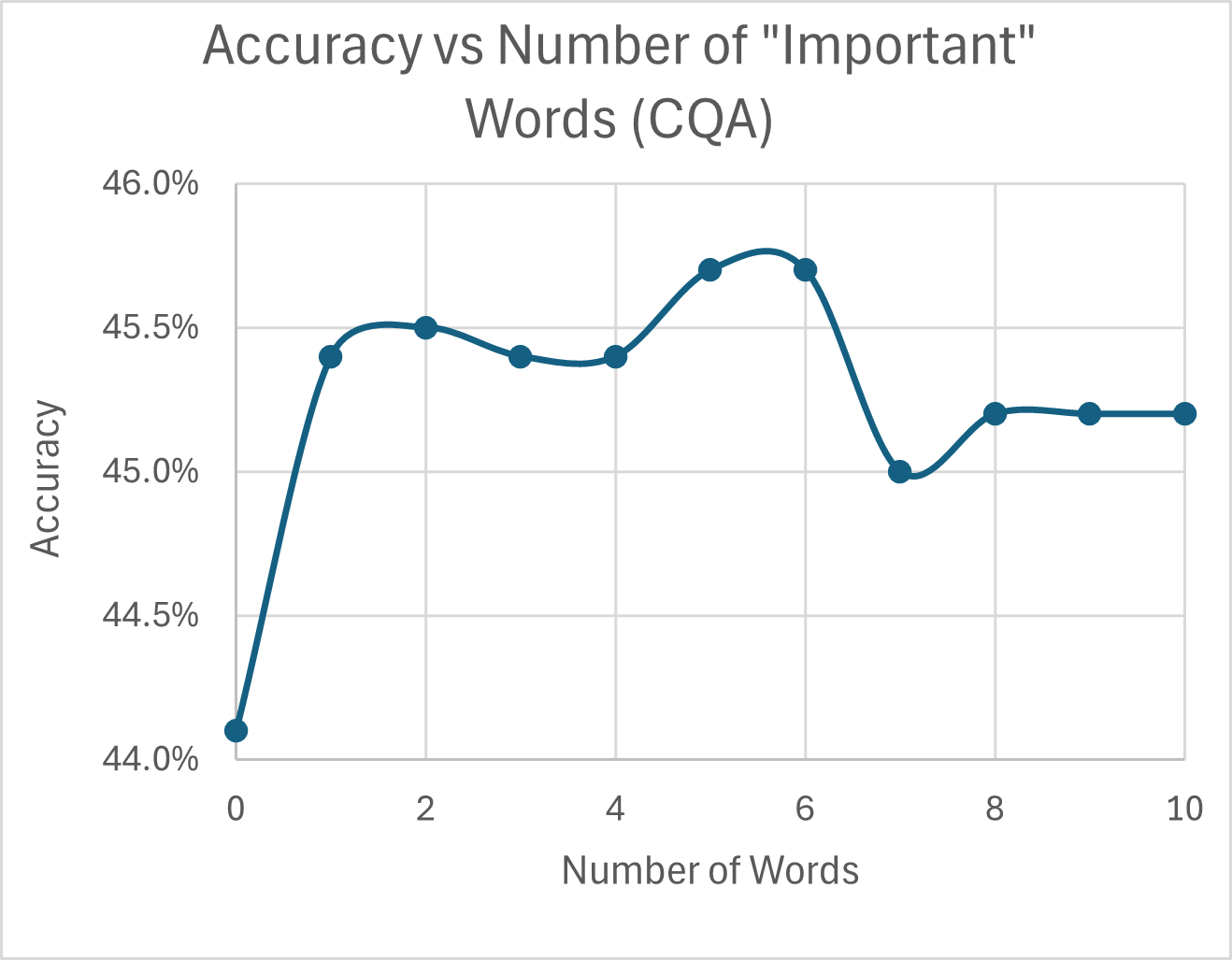}

\end{minipage}

\caption{The figure shows the performance of the model versus the number of tokens provided as rationales}
\label{fig2}
\end{figure}

Figure \ref{fig2} illustrates the outcomes of these experiments. It shows that improvement over the baseline standard fine-tuning begins with the inclusion of just one word extracted from the teacher model as a rationale. This enhancement continues to increase as we include up to four words. However, when using between 4 to 6 words, the accuracy experiences fluctuations, and beyond six words, there is a noticeable drop in performance.

From these observations, we conclude that the model can indeed benefit from incorporating as few as one word and up to ten words as rationales. Nevertheless, the optimal number of words, or the ``sweet spot", appears to be around five. Therefore, in our experiments, we adopted five as the standard number of important words to use as rationales.

The implication of these results is multifaceted. First, it confirms that strategic ``hints" from a larger teacher model can enhance the learning efficiency of a smaller student model, mirroring a focused guidance approach. Too much information, conversely, can overwhelm the model, akin to information overload in humans, which can impede rather than facilitate understanding. Moreover, the consideration of context, such as the input length in datasets like CQA, which may feature inputs as brief as twelve words, further supports the rationale behind limiting hints to five words. Providing nearly the entire input as a hint can dilute the effectiveness of the distillation process. 

\subsection{Random Words as Rationales}

In this section, to verify that the improvement in our model's performance is not due to random factors, we conducted a controlled experiment. Instead of feeding the model with the most important tokens, which are identified based on the highest attribution scores, we provided it with five random words from the input. This test aimed to assess the impact of using non-targeted words compared to strategically selected ones. As anticipated and detailed in table \ref{tab3}, the introduction of random words instead of the most significant tokens led to a decline in the model's performance on the CQA dataset. This result underscores the importance of selecting appropriate, influential words for enhancing model effectiveness.

\begin{table}[htbp]
\caption{The table displays the accuracy of fine-tuning T5-small under three conditions: without using any words as rationales, with words extracted from flan-t5-3b used as rationales, and with random words used as rationales.}\label{tab3}
\centering
\begin{tabular}{|c|c|}
\hline
\textbf{Teacher} &   \textbf{Accuracy}\\
\hline
No Teacher & 44.1\% \\
Flan-T5-3b & 45.7\%  \\
Random words &  42.1\%  \\
\hline
\end{tabular}
\end{table}

\section{Discussion}
\subsection{Token Analysis Efficacy with the Teacher Model}
We further explored our method by analyzing the words extracted from the teacher model, aiming to understand why it is effective. This investigation focused on the CQA dataset, which is well-suited for this analysis because it consists of multiple-choice questions with answers provided in the input. Our approach involved comparing the top-5 important tokens extracted by the teacher model with the ground truth label. For example, consider a dataset input: “A boss may like an employee’s ambition, so the employee may get put what? Choices are (a) in charge of project (b) conquer opponent (c) go to school (d) begin work (e) webisode,” where the correct answer is (a) in charge of project. We then examined whether any of the top-5 important tokens, are part of the ground truth  such as ``charge" or ``project" in this case. Our findings indicate that in around 68\% of cases, the correct answer, or a part of it, is among these top 5 most influential tokens. Furthermore, we observed that even when the correct answer is not directly among the important tokens, these tokens are still contextually related. For instance, if the answer is ``monkey", the important tokens might include ``banana". Similarly, if the answer is ``New York City", the important tokens could be ``empire", ``state", ``location", and for ``hotel", they might include ``book", ``room". This pattern suggests that the extracted tokens, even when not directly answering the question, are relevant and contribute to understanding the context of the question.

\subsection{Similarity between Compared Models} 

In this section, we aim to gain deeper insights into the differences between two methods by comparing the predictions made by the distillation step-by-step model and our model. Our investigation focuses on two scenarios: instances where both models correctly predict the answer, and instances where both models make an incorrect prediction, specifically examining whether the incorrect predictions are similar.

Initially, we assess the proportion of instances where both models correctly predict the answers. The results, presented in table \ref{tab5}, indicate a high degree of overlap in correct predictions across three datasets. For instance, in the ANLI dataset, 95.2\% of the answers correctly predicted are the same for both models. This suggests that when the distillation step-by-step model accurately predicts an answer, there is a 95.2\% likelihood that our model will also predict correctly.

\begin{table}[htbp]
\caption{Table shows the percentage of overlapping correct predictions by the distillation Step-by-Step model and our model across different datasets.}\label{tab5}
\centering
\begin{tabular}{|c|c|c|}
\hline
\textbf{Dataset} & \textbf{Random Chance} &  \textbf{Similarity}\\
\hline
ANLI & 33.3\% & 95.2\% \\
e-SNLI & 33.3\% & 95.5\%  \\
CQA & 20.0\% &  85.1\%  \\
SVAMP & -- &  81.1\%  \\
\hline
\end{tabular}
\end{table}

Conversely, table \ref{tab6} reveals that the models also have a high likelihood of making the same incorrect prediction. Specifically, in the ANLI dataset, where the task involves a three-way classification problem (entailment, contradiction, and neutral), there is a 91.1\% chance that if the distillation step-by-step model incorrectly predicts an entailment relation between the hypothesis and the premise, our model will also incorrectly predict the same.

The high overlap in correct predictions between the models underscores their capability to similarly interpret and analyze the data, reflecting their effectiveness in learning from the training set. On the other hand, the significant concordance in incorrect predictions highlights potential systematic errors or biases that both models may share, possibly due to similarities in their training data or inherent model architectures.

\begin{table}[htbp]
\caption{Table shows the percentage of similar incorrect predictions between the distillation Step-by-Step model and our model across various datasets.}\label{tab6}
\centering
\begin{tabular}{|c|c|c|}
\hline
\textbf{Dataset} &   \textbf{Random Chance} &   \textbf{Similarity}\\
\hline
ANLI & 33.3\% &  91.1\% \\
e-SNLI & 33.3\% & 90.8\%  \\
CQA & 20.0\% &  90.4\%  \\
SVAMP & -- &  87.2\%  \\
\hline
\end{tabular}
\end{table}

\subsection{Integrated Gradients Method}
In our research, we also explored an advanced gradient attribution method known as Integrated Gradients (IG) \cite{ig}, which offers a more sophisticated analysis of feature importance in deep neural networks. Integrated Gradients overcomes several limitations found in simpler attribution methods, such as gradient saturation and insensitivity to certain model behaviors. It achieves this by integrating the gradients of the model's output relative to its inputs along a straight-line path from a chosen baseline (typically a non-informative input) to the actual input.

However, implementing this method is computationally intensive. Consequently, we applied it to the SVAMP dataset, which has a relatively small number of training samples (only 800). The results from this experiment were not as promising as we had hoped. We observed no improvement in the performance of the t5-small model using Integrated Gradients compared to simpler attribution methods. The accuracy remained at 50.0\% on the SVAMP dataset. This finding suggests that using a more advanced method like Integrated Gradients to extract the top-k important tokens does not necessarily confer any advantage in this context.  

\section{Limitation}

Our research demonstrates an effective and straightforward approach for generating rationales from a large model to enhance a smaller model. However, a significant performance gap between the teacher model (the large model) and the student model (the small model) still exists. As illustrated in table \ref{tab4}, the flan-t5-3b, which serves as the teacher model, significantly outperforms the smaller t5-small model across all four datasets we tested.

This limitation underscores the inherent challenge in knowledge distillation and model scaling. While our method effectively transfers knowledge from a large to a small model, the reduced capacity of the smaller model limits its ability to fully replicate the performance of its larger counterpart. This performance discrepancy highlights the trade-offs involved in model downsizing, where gains in efficiency and deployability often come at the cost of reduced accuracy and overall capability.

Addressing this gap remains a key area for future research. Efforts could focus on developing more advanced distillation techniques or optimizing small models to better capture and utilize the knowledge transferred from larger models. Our findings lay the groundwork for such exploration, offering insights into the dynamics of knowledge transfer between models of varying sizes and complexities.

\begin{table}
\caption{The table compares the performances of the teacher model and the distilled student models. }\label{tab4}
\centering
\begin{tabular}{|c|c|c|c|c|c|c|}
\hline
\textbf{Model} &  \textbf{Base Model}  &  \textbf{ \# Param.} & \textbf{ANLI} &  \textbf{e-SNLI} &  \textbf{CQA} &  \textbf{SVAMP} \\
\hline
Dist. step-by-step & T5-small & 60M & 39.5  & 84.6  & 44.4 & 49.5 \\
Ours & T5-small & 60M &  39.5  &  84.8 &  45.7  &  50.0  \\
\hline
FLAN-T5-3b & FLAN-T5-3b & 2.7B &  \textbf{67.3}  & \textbf{93.1}  &  \textbf{81.4}  &  \textbf{86.0}  \\
\hline
\end{tabular}
\end{table}

\section{Conclusion and Outlook}

In this work, we demonstrated an effective approach to distilling knowledge into student models by calculating the attribution scores of the teacher model's input and using them as rationales for the student model. This method is based on the theory that a high gradient value for a token in the input indicates that minor changes in that token can significantly impact the model’s prediction, emphasizing its importance in influencing the model's output. Conversely, a low gradient value suggests that the token has minimal effect on the prediction, signifying its lesser importance.

We tested our method on four diverse datasets, which included tasks in natural language inference, commonsense question answering, and arithmetic word problems. The results showed that using t5-small as the student model, our method enhances performance across these datasets. Additionally, we explored why this method is effective by analyzing the top important tokens used as rationales. Particularly in the multiple-choice CQA dataset, where the answer is included in the input, we found that the top-5 extracted tokens contain the answer or a part of it 68\% of the time.

While our method offers a simple and effective means of knowledge distillation, there is still a significant performance gap between the teacher and student models, with the teacher model outperforming the student by a large margin. We hope that this work contributes to the field of knowledge distillation and inspires further research to narrow the gap between large and small models. This endeavor is crucial for advancing the efficiency and applicability of language models in various real-world scenarios.

\section*{Acknowledgements}

This work was funded by the Deutsche Forschungsgemeinschaft (DFG, German Research Foundation). The cluster used to train the models was also funded by the German Research Foundation (DFG) - 456666331.
%
%

%
%
%

\bibliographystyle{splncs04}
\bibliography{ref.bib}

\end{document}